\documentclass[preprint,3p]{elsarticle}
\usepackage[utf8]{inputenc}
\usepackage[T1]{fontenc}
\usepackage{amsthm,amsmath,amssymb,amsfonts}
\usepackage{rotating}
\usepackage{tikz}
\usepackage[inline]{enumitem}
\usepackage{todonotes}
\setuptodonotes{backgroundcolor=white}
\usepackage{subfig}
\usepackage{pifont}
\usepackage[hidelinks]{hyperref}
\usepackage[capitalise,noabbrev,sort&compress]{cleveref}
\usepackage{natbib}   
\setcitestyle{square,comma,numbers,sort&compress}
\usepackage{xspace}  
\usepackage{tikz}
\usetikzlibrary{calc,positioning,shapes.geometric,shapes.misc, positioning}
\usepackage{varwidth}

\newcommand{\DefectOnt}{\emph{DefectOnt}\xspace}

\begin{document}

\begin{frontmatter}
\title{An Ontology for Defect Detection \\ in Metal Additive Manufacturing}
\author[label1]{Massimo Carraturo}
\author[label2]{Andrea Mazzullo}
\address[label1]{Department of Civil Engineering and Architecture, University of Pavia, Italy}
%
\address[label2]{KRDB Research Centre, Faculty of Computer Science, Free University of Bozen-Bolzano, Italy}


\begin{abstract}
A key challenge for Industry 4.0 applications is to develop control systems for automated manufacturing services that are capable of addressing both data integration and semantic interoperability issues, as well as monitoring
and decision making tasks.
To address such an issue in advanced manufacturing systems, principled knowledge representation approaches based on formal ontologies have been proposed as a foundation to information management and maintenance in presence of heterogeneous data sources.
In addition, ontologies provide reasoning and querying capabilities to aid domain experts and end users in the context of constraint validation and decision making.
Finally, ontology-based approaches to advanced manufacturing services can support the explainability and interpretability of the behaviour of monitoring, control, and simulation systems that are based on black-box machine learning algorithms.
In this work, we provide a novel ontology for the classification of process-induced defects known from the metal additive manufacturing literature.
Together with a formal representation of the characterising features and sources of defects, we integrate our knowledge base with state-of-the-art ontologies in the field.
Our knowledge base aims at enhancing the modelling capabilities of additive manufacturing ontologies by adding further defect analysis terminology and diagnostic inference features.

\end{abstract}


\end{frontmatter}



\section{Introduction}\label{sec:intro}

\noindent
Fully automated manufacturing systems,
as well as
production-as-a-service frameworks, represent a cornerstone of Industry 4.0 applications~\cite{gaiardelli2021product}.
In this context, additive manufacturing (AM), and specifically \emph{metal additive manufacturing} (MAM), is
particularly suited to
industrial
paradigms based on automation, flexibility, and efficiency.
Indeed, MAM can be considered as a native digital technology, providing a seamless workflow from the digital design environment to the final product, which can be potentially completed without any human intervention~\cite{Gibson2015}.

However, a broader adoption of MAM technologies in industry is still hindered by such factors as:
\begin{enumerate*}[label=$(\roman*)$]
\item lack of widely adopted standardisations and specifications of material properties, machines, and processes~\cite{king2015laser};
	\item lack of adequate digital infrastructures, and interoperability issues between different production environments~\cite{badiru2017additive};
	\item lack of
accessible interfaces
providing
process information that is easily interpretable by non-experts~\cite{lawrence2017advances};
	\item lack of advanced control systems capable of automatically adjusting, at run-time, the production parameters~\cite{mani2015};
	\item challenges in quality assurance due part accuracy and variability~\cite{leary2019design}.
\end{enumerate*}

Thus, achieving semantically transparent and interoperable data sets and systems, to address Points~$(i)$, $(ii)$ and $(iii)$ above, is  arguably
of paramount importance.
%
In this direction, several approaches based on ontology engineering and knowledge representation techniques have been proposed~\cite{garetti2012role,belkadi2016towards,SanEtAl19,SanEtAl19a,preisig2020ontologies}.
%
Broadly conceived as formal specifications of conceptualisations over a domain of interest, computational ontologies (cf.~\cite{GuaEtAl09} and references therein) have been in particular investigated as a tool to improve interoperability of additive manufacturing systems that involve human-intensive and domain-expert knowledge management tasks (cf.~Section~\ref{sec:state-of-the-art} for a literature survey).
%

To the best of our knowledge, however, despite the number of domain-specific ontologies proposed in the literature to address interoperability issues, less attention has been devoted to another crucial aspect of AM applications, in particular of \emph{Powder Bed Fusion} (PBF) MAM: that of \emph{defect diagnosis} and \emph{correction}.
PBF is a layer-by-layer process, where a layer of metal powder is spread by means of a roller on top of a build plate, and metal powder particles are selectively melted by means of a localised moving laser heat source \cite{frazier2014metal}.
At industrial level, PBF MAM is a widespread technology, due to its capability to deliver parts with high surface quality and remarkable mechanical properties.
Nonetheless, the localised nature of the
melt pool induces rapid melting-solidification cycles that are responsible for part deflections and residual stresses \cite{Bartlett2019}. 
Moreover, due to their multi-scale and multi-physics nature,
phenomena involved in PBF processes are difficult to control.
Finally, the overall process is characterised by complex and not yet fully understood relationships among material microstructure, part geometry, process parameters, and mechanical properties and performances \cite{smith2016linking}.
Such issues can lead to process-induced material discontinuities, e.g. lack-of-fusion and keyhole porosity, balling, crack and delamination \cite{DUPLESSIS2020108385}.

Even if these process-related defects play a key role influencing part properties and performances (e.g., elastic and elastoplastic behaviour, ultimate tensile stress, and fatigue life), an ontology-based representation of MAM defects, together with their main properties surveyed in the literature~\cite{bian2017laser}, is not yet available.
Such a principled approach, integrating observational data with formalised domain knowledge to determine the causality links and the complex process-structure-property relationships in manufacturing processes, represents an important preliminary step for the development of reliable monitoring and control systems capable of addressing Points~$(iv)$ and~$(v)$ above~\cite{debroy2018additive}.
 %
%
%

Our contribution aims at filling this gap, by introducing the novel \DefectOnt \footnote{\href{https://github.com/AndreaMazzullo/DefectOnt}{\texttt{https://github.com/AndreaMazzullo/DefectOnt}.}} ontology for MAM, specifically PBF-based, defects.
This ontology relies on a modular structure, and it aligns with other upper and domain-specific ontologies from the literature, to favour development, interoperability and maintenance.
It includes axioms covering the following dimensions:
\begin{enumerate*}[label=$(\roman*)$]
\item MAM-based categories of defects and related properties;
\item spatial notions to express geometrical and topological characteristics;
\item dimensional characteristics requiring a vocabulary of metrological terms;
\item sensor-related concepts for observational properties.
\end{enumerate*}
\DefectOnt is implemented in the OWL 2~\cite{GraEtAl08}\footnote{\href{https://www.w3.org/TR/owl2-overview/}{\texttt{https://www.w3.org/TR/owl2-overview/}.}}, using the open-source Prot\'{e}g\'{e} ontology editor~\cite{KnuEtAl04}\footnote{\href{https://protege.stanford.edu/}{\texttt{https://protege.stanford.edu/}.}}.
%


The present article is organised as follow. In Section~\ref{sec:state-of-the-art}, we discuss related work on MAM defects and AM ontologies. Then, in Section~\ref{sec:ont-design-meth}, the design methodology and the development phases of our MAM defect ontology are described. In Section~\ref{sec:AM-defect-ontology},  we illustrate the \emph{DefectOnt} ontology, presenting in detail the modules that constitute it. Finally, in Section~\ref{sec:conclusions}, we discuss future research directions and the main conclusions of the present work.

\section{Related work}\label{sec:state-of-the-art}

\subsection{MAM-related literature}

\noindent
\citet{malekipour2018common}  identify, analyze, and classify the most common defects in PBF MAM, defining the relationships among defects and their contributing parameters. To develop a suitable online monitoring control strategy, defects are organized into categories based on their manufacturing features and control purposes.
\citet{kyogoku2020review} review the literature regarding defect generation mechanisms in PBF processes and their mitigation strategies. 
\citet{snow2020invited} outline the state-of-the-art knowledge of gas porosity and lack-of-fusion flaws due to melt pool instabilities in PBF processes.
\citet{grasso2017process} present a defect classification of PBF process-induced defects based on process signatures, to support in-situ monitoring and online defect detection.
The present contribution follows defects taxonomies proposed in \cite{grasso2017process} (and references therein), integrating it with other literature resources (of both non-ontological and ontological nature).

%
\subsection{Ontology-related literature}
\noindent
Given its closely related focus (despite not overlapping with ours,  content-wise), we relied on \emph{ExtruOnt}, an ontology for the description of an extruder components proposed by~\citet{RamEtAl20}, as a gold standard for the development of our knowledge base, adhering to the authors' methodological, design, and presentation choices. 

With a broader scope, other upper or domain ontologies for AM have been proposed in the literature.
The Manufacturing's Semantics Ontology (MASON)~\cite{LemEtAl06} is an upper ontology for the conceptualisation of core additive manufacturing domain notions.
The US~National Institute of Standards and Technology (NIST) propose an ontology for AM, that we label NIST AM~\cite{WitEtAl14,WitEtAl16}, to support the development of laser and thermal metamodels.
%
Towards interoperable knowledge and data management in applications, \citet{SanEtAl19} introduce another ontology for AM (Onto4Additive), based on the upper Descriptive Ontology for Linguistic and Cognitive Engineering (DOLCE)~\cite{BorMas09}.
%

Other related AM ontologies include the following (cf. also literature reviews in~\citet{SanEtAl19} and~\citet{RamEtAl20}):
the Additive Manufacturing Ontology (AMO)~\cite{AliEtAl19},
based on the upper Basic Formal Ontology (BFO)~\cite{arp2015building};
the Innovative Capabilities of Additive Manufacturing
(ICAM) ontology~\cite{HagEtAl18};
%
a Smart Applications Reference (SAREF)~\cite{appliances2017smartm2m}
extension for semantic interoperability in the industry and manufacturing domain (SAREF4INMA)~\cite{DanEtAl18};
the Semantically Integrated Manufacturing Planning Model (SIMPM)~\cite{vsormaz2019simpm};
the Manufacturing Resource Capability Ontology (MaRCO)~\cite{JarEtAl19};
the Manufacturing Service Description Language (MSDL) ontology~\cite{ameri2006upper};
the
Politecnico di Milano--Production Systems Ontology (P-PSO) ~\cite{garetti2012p};
the
Ontology of Standard for the Exchange of Product model data (OntoSTEP)~\cite{BarEtAl12};
the ontologies proposed by~\citet{DinRos17},~\citet{liang2018ontology},~\citet{roh2021ontology}, and~\citet{li2022description}.
%
Finally, within the EU-funded project EMMC, the recently proposed European Materials Modelling Ontology (EMMO)~\cite{HorEtAl20} provides a standard representational upper (nominalistic) ontology framework based on state-of-the-art knowledge on material modelling and engineering.

\section{Ontology design and development}
\label{sec:ont-design-meth}
\noindent
To develop our ontology, we followed the approach proposed by~\citet{RamEtAl20}, with the adoption of the NeOn Methodology~\cite{SuaEtAl12} and in particular of the Six-Phase + Merging Phase Waterfall Ontology Network Life Cycle Model.
This model, allowing for a flexible interplay between pre-existing ontological and non-ontological resources, consists of the following phases, which will be detailed in the remainder of this section:
initiation,
reuse,
merging,
re-engineering,
design,
implementation,
and
maintenance.

\subsection{Initiation phase}
\noindent
As by the methodological framework of~\citet{SuaEtAl09},
we initially developed an Ontology Requirements Specification Document (ORSD), summarised in Table~\ref{tab:orsd}, with the following goals:
defining the purpose and the scope of our ontology; selecting the implementation languages; identifying intended users and uses of the ontology; formulating in natural language groups of competency questions (CQs), to be expressed and answered by our ontology;
providing a pre-glossary of terms appearing in the CQs.

To better illustrate the purpose and the intended uses of our ontology, we present the following simplified scenario, involving a MAM production service monitored and regulated by a control system relying, for instance, on a machine learning architecture.
Suppose that, during a 3D printing process, the monitoring system detects a feature that is classified (by means of, e.g., a pre-trained convolutional neural network) as a porosity defect, consisting of a void encapsulated within bulk material.
Moreover, assume that the system controller (based on, e.g., a reinforcement learning mechanism), in an attempt to mitigate the propagation of the feature, sequentially modifies relevant build chamber parameters, powder handling and deposition system parameters, and the laser scanning speed, observing that only the latter has an impact on limiting the defect.
The purpose of our ontology is to formalise the domain knowledge required to infer that the detected porosity is an instance of a process-induced defect, rather than of an equipment-induced one (given that all other possible equipment-related parameters have been ruled out as influences on the feature).

In such contexts, our ontology can be used to:
provide a (both machine- and human-readable) structured representation of MAM defects, including their main characteristics and mutual relationships;
enrich the online monitoring and troubleshooting capabilities of controllers by means of logic-based reasoning services;
improve the user interface to MAM production processes, integrating black-box controller systems with a user-queryable and explainable diagnostic framework.

Finally, we have grouped the CQs along four dimensions, constituting the backbone of our ontology modular structure:
\begin{enumerate*}[label=$(\roman*)$]
\item MAM-related (CQAm);
\item spatial-related (CQSp);
\item measure-related (CQMe);
and
\item sensor-related (CQSe).
\end{enumerate*}

\setlist[description]{font=\normalfont\itshape}

\begin{table}
\centering
\caption{Summary of \emph{DefectOnt} Ontology Requirements Specification Document.}
\label{tab:orsd}
\begin{tabular}{ p{\textwidth} }
 \hline
{\footnotesize{
\begin{description}
	%
	\item[{Purpose}] Representation of and reasoning about defects and defect sources in MAM processes, enhancing inference-based diagnostic capabilities to support the user in production control and decision making phases.
	\item[{Scope}]
	Diagnosis and troubleshooting in MAM machines and processes. Potential application contexts are research laboratories and industrial settings.
	\item[{Implementation languages}] 
	OWL 2 (description logic syntax).
	\item[{Intended users}]
	Domain experts, product designers, company employees.
	\item[{Intended uses}]
	\begin{itemize}	[topsep=0pt,before=\leavevmode\vspace{-1.5em}]
		\item[]
		\item Aid user understanding of the classification of MAM defects and corresponding sources.
		\item Enhance automated MAM process control systems with high-level reasoning capabilities.
		\item Guide user interaction with automated MAM process control systems by means of ontology-based explainable diagnostic services.
	\end{itemize}
	\item[{Ontology requirements}]
		\begin{itemize}[topsep=0pt,before=\leavevmode\vspace{-1.5em}]
			\item[]
			\item Non-functional requirements:
			Ontology based on defect classifications from state-of-the-art surveys in the MAM literature.
			%
			\item Functional requirements: Groups of competency questions (CQs).
			\begin{itemize}
				\item[CQAm] MAM-related CQs
				\begin{enumerate}[label=CQAm.\arabic*]
					\item Is the feature ${\sf d}$ an instance of a porosity defect?
					\item Is defect ${\sf d}$ induced by a process parameter?
					\item \ldots
				\end{enumerate}
				\item[CQSp] Spatial-related CQs
				\begin{enumerate}[label=CQSp.\arabic*]
					\item Is the MAM product {\sf pr} affected by any equipment-induced surface defects?
					\item Are there balling defects located on the surface layers of product {\sf pr}?
					\item \ldots
				\end{enumerate}
				\item[CQMe] Measure-related CQs
					\begin{enumerate}[label=CQMe.\arabic*]
						\item What is the melt pool temperature in degrees Celsius?
	\item What is the thickness in millimetres of layer {\sf l}?
	\item What is the length in metres of the cracking defect instance {\sf d}?
	\item \ldots
					\end{enumerate}
				\item[CQSe] Sensor-related CQs
				\begin{enumerate}
					\item[CQSe.1] Which sensors are hosted by platform {\sf pl}?
					\item[CQSe.2] Does any sensor hosted by platform {\sf pl} observe an instance of a porosity defect?
						\item[CQSe.3] \ldots
				\end{enumerate}
			\end{itemize}
		\end{itemize}
	\item[Pre-glossary of terms] Defect, porosity, cracking, balling, equipment-induced defect, process-induced defect, product, layer, sensor, platform, melt pool, \ldots
\end{description}
}}\\
\hline
\end{tabular}
\end{table}

\subsection{Reuse phase}
\noindent
%
In order to obtain the domain knowledge required to express and formalise relevant properties of MAM defects, we collect material from both ontological and non-ontological resources. While the latter are based on (not yet formalised) literature on MAM defects, we relied on already existing additive manufacturing ontologies to gather structured knowledge related to MAM processes.
The selected ontologies were preferred over other available resources based on the following criteria:
(i) possibility of integrating class and property hierarchies with other non-ontological resources; and (ii) vocabularies capable of expressing properties of defects determined by the CQs.
In the following, we present the relevant material divided along the four dimensions determined by the groups of CQs.

\begin{description}[font=\normalfont\itshape]
	\item[MAM-related resources]
	As our main non-ontological resources related to MAM defects, we identify:~\citet{grasso2017process}, providing a classification that relates each kind of defect with the main causes analysed in the literature;
	\citet{malekipour2018common}, similar to the previous article in scope and purpose, but providing a different taxonomy of defects;
	\citet{snow2020invited}, focusing mainly on the internal porosity classification.
	To exploit formal representations of the main notions involved in the MAM domain, while maintaining a neutral and interoperable environment, we use instead the following ontologies:
	NIST AM~\cite{WitEtAl16}\footnote{\href{https://github.com/iassouroko/AMontology}{\texttt{https://github.com/iassouroko/AMontology}.}};
	and
	Onto4additive~\cite{SanEtAl19}\footnote{\href{https://ontohub.org/repositories/additive-manufacturing}{\texttt{https://ontohub.org/repositories/additive-manufacturing}.}}.
			Finally, as a methodological guidance and as a source for specific module development (cf.~\emph{Measure-related resources}),
			we rely on the \emph{ExtruOnt} ontology~\cite{RamEtAl20}\footnote{\href{http://siul02.si.ehu.es/bdi/ontologies/ExtruOnt/docs/}{\texttt{http://siul02.si.ehu.es/bdi/ontologies/ExtruOnt/docs/}.}}.

	\item[Spatial-related resources] 
	To model spatial-related, geometrical or topological, concepts and relations, we choose GeoSPARQL 1.1~\cite{CarEtAl22}\footnote{\href{https://opengeospatial.github.io/ogc-geosparql/geosparql11/index.html}{\texttt{https://opengeospatial.github.io/ogc-geosparql/geosparql11/index.html}.}}. Additional spatial-related concepts are inherited from the MASON~\cite{LemEtAl06}\footnote{\href{https://sourceforge.net/projects/mason-onto/}{\texttt{https://sourceforge.net/projects/mason-onto/}.}} ontology.

	\item[Measure-related resources]\label{item:measureresource} To express metrological features in our ontology, we rely directly on the \emph{OM4ExtruOnt} module developed by~\citet{RamEtAl20}. This module is obtained by removing all the classes and properties not relevant to the manufacturing setting from the OM ontology~\cite{RijEtAl13}\footnote{\href{https://github.com/HajoRijgersberg/OM}{\texttt{https://github.com/HajoRijgersberg/OM}.}}.

	\item[Sensor-related resources]
	As main ontological resource to conceptualise observation- and sensor-related notions, we select the Semantic Sensor Network (SSN) ontology~\cite{HalEtAl19}\footnote{\href{https://www.w3.org/TR/vocab-ssn/}{\texttt{https://www.w3.org/TR/vocab-ssn/}.}}. 
	
\end{description}

\subsection{Merging phase}
\noindent
%
With the aim of improving semantic interoperability and knowledge exchange in applications, we structure the upper or domain ontologies identified in the reuse phase so to be aligned with the \DefectOnt framework.
In order to merge these ontologies, we perform the following steps.
\begin{enumerate}[label=$(\roman*)$]
	\item We select higher-level classes (with respect to our defect-related concepts) to be introduced in the \DefectOnt hierarchy (e.g., {\sf Entity} from MASON ontology,  {\sf Abstract}, {\sf Physical}, {\sf Characteristic} from NIST AM ontology, {\sf PhysicalObject} from Onto4Additive ontology, {\sf SpatialObject} from GeoSPARQL ontology, {\sf Property} from SSN ontology).
	\item We identify classes that (for our modelling purposes) formalise interchangeable notions, in order to set them as equivalent in the class hierarchy of the \DefectOnt ontology (e.g., {\sf PhysicalObject} from the Onto4Additive and {\sf SpatialObject} from the GeoSPARQL; {\sf Characteristic} from NIST and {\sf Property} from SSN).
	\item We devise class sub-hierarchies to connect the vocabulary and the axioms from different ontologies (e.g. {\sf Characteristic} from NIST and {\sf ObservableCharacteristic} from SSN introduced as subclass; {\sf Physical} from NIST and {\sf SpatialObject} from GeoSPARQL introduced as subclass).
\end{enumerate}

\subsection{Re-engineering phase}
\noindent
As already mentioned, our main non-ontological resources identified in the reuse phase are~\citet{grasso2017process} (in particular, Table~1 and references therein), \citet{snow2020invited}, and \citet{malekipour2018common}.
These references are used to extract a conceptual framework based on defect types and corresponding sources, compatible with the purpose, the intended uses, and the implementation languages of the \DefectOnt ontology.

\subsection{Design phase}
\noindent
To facilitate development, interoperability, and maintenance (in line with the approach by~\citet{RamEtAl20}), our \emph{DefectOnt} ontology is based on a modular structure, reflecting the dimensions that emerge with the ORSD development in the initiation phase.
The \DefectOnt ontology consists of the following three modules.
\begin{enumerate}[label={\arabic*.}]
	\item \emph{MAM4DefectOnt}, a MAM-related module importing the following ontologies:
			\begin{enumerate}[label*=\arabic*]
			\item \emph{NIST4DefectOnt}, a module obtained from the NIST AM ontology;
			\item \emph{ONTO4ADD4DefectOnt}, derived from the \emph{Onto4Additive} ontology;
			\item \emph{Spatial4DefectOnt}, to represent spatial-related notions, in turn consisting of the following modules:
			\begin{enumerate}[label*=.\arabic*]
				\item \emph{GeoSPARQL4DefectOnt}, obtained from the GeoSPARQL1.1 ontology;
				\item \emph{MASON4DefectOnt}, derived from the MASON ontology.
			\end{enumerate}
		\end{enumerate}
	\item \emph{Measure4DefectOnt}, a measuring-related module importing \emph{OM4ExtruOnt}, a submodule originally developed for the \emph{ExtruOnt} ontology from the OM ontology.
	\item \emph{Sensor4DefectOnt}, a sensor-related module consisting of the submodule \emph{SSN4DefectOnt} obtained from the SSN ontology.
\end{enumerate}

To design our ontology modules, we perform the following main steps:

\begin{enumerate}[label=$(\roman*)$]
	\item\label{item:one} In a preliminary class hierarchy of the \emph{MAM4DefectOnt} module, we structure the domain knowledge formalised from the non-ontological resources during the re-engineering phase (e.g., by introducing the {\sf Defect} class and related sub-classes).
	\item\label{item:two} We prune the class hierarchies of the ontologies selected in the re-use phase, maintaining only the most relevant vocabulary and axioms,
		to obtain the submodules for \emph{MAM4DefectOnt}, as well as for its submodule \emph{Spatial4DefectOnt}, \emph{Measure4DefectOnt}, and \emph{Sensor4DefectOnt}.
	\item\label{item:three} We refine the ontology formalisation of Step~\ref{item:one} by adding to the \emph{MAM4DefectOnt} module axioms based on classes and properties from submodules devised in Step~\ref{item:two} (e.g., {\sf Defect}-related axioms based on the {\sf influencedBy} property from NIST ontology).
	\item We enrich the class hierarchies of Step~\ref{item:two} submodules to further improve the ontology formalisation process (e.g., introducing in \emph{Spatial4DefectOnto} the class {\sf Ball} as subclass of {\sf Geometric\_entity} from MASON ontology).
	\item We modify class hierarchies and axioms from the submodules obtained in Step~\ref{item:two} to adhere to the non-ontological resource formalisation performed in Steps~\ref{item:one} and~\ref{item:three} (e.g., by setting {\sf EquipmentParameter} as a sibling class of {\sf ProcessParameter}, rather than subclass (as in NIST), in accordance with~\citet{grasso2017process}).
	\item We perform a terminological normalisation process, to align the naming scheme for short IRIs of \DefectOnt classes (e.g., by changing {\sf influencedBy} from NIST to {\sf isInfluencedBy}; {\sf Geometry\_entity} from MASON to {\sf GeometricEntity}).
\end{enumerate}

\subsection{Implementation phase}
\noindent
Initially partly expressed by means of a description logic (DL) language~\cite{BaaEtAl17}, our ontology is fully implemented in OWL 2, using the Prot\'{e}g\'{e} ontology editor.
According to the Prot\'{e}g\'{e} ontology metrics, \DefectOnt can be expressed with the logical constructs provided by the $\mathcal{ALCHOIQ}(\mathcal{D})$ DL language.

\subsection{Maintenance phase}
%
\noindent
As a time-consuming and error-prone task, the knowledge extraction process for the design of the \DefectOnt ontology comes with a (still active) maintenance protocol. As prescribed by the selected Ontology Network Life Cycle Model, whenever a modelling error is detected, our ontology re-enters one among the reuse, merging, re-engineering, or design phases for corrections.

\section{Ontology modules}
\label{sec:AM-defect-ontology}

\begin{figure}[t]
\centering
\begin{tikzpicture}[scale=0.5, every node/.style={scale=0.5},
    >=stealth,
    box/.style = {text width=2cm, font=\small},
    node distance=3cm,
    database/.style={
      cylinder,
      shape border rotate=90,
      aspect=0.25,
      draw
    }
  ]
    \node[database] (defectont) at (0,0) {\shortstack{\DefectOnt\\Ontology}};

    \node[database] (mam) at (0,-3) {\shortstack{\emph{MAM}\\\emph{4DefectOnt}}};

        \node[database, align=center] (nist) at (-4.5,-4.5)
        {
        \emph{NIST} \\ \emph{4DefectOnt}
        };
		\node[database, align=center] (onto4add) at (4.5,-4.5) 
		{
		\emph{ONTO4ADD} \\ \emph{4DefectOnt}
		};   
		 
    \node[database] (spatial) at (0,-6) {\shortstack{\emph{Spatial}\\\emph{4DefectOnt}}};
        \node[database, align=center] (geosparql) at (-4,-8) {\emph{GeoSPARQL} \\ \emph{4DefectOnt}};
        \node[database, align=center] (mason2) at (4,-8) {\emph{MASON} \\ \emph{4DefectOnt}};
    %
    \draw[->] (spatial) -- (geosparql);
    \draw[->] (spatial) -- (mason2);

    \draw[->] (defectont) -- (mam);
    \draw[->] (mam) -- (nist);
    \draw[->] (mam) -- (spatial);
    \draw[->] (mam) -- (onto4add);
    
    \node[database, align=center] (measure) at (-3,3) {\emph{Measure}\\\emph{4DefectOnt}};
        \node[database, align=center] (om) at (-6,6) {\emph{OM} \\ \emph{4ExtruOnt}};
    \draw[->] (defectont) -- (measure);
    \draw[->] (measure) -- (om);

    \node[database, align=center] (sensor) at (3,3) {\emph{Sensor}\\\emph{4DefectOnt}};
    \node[database, align=center] (ssn) at (6,6) {\emph{SSN} \\ \emph{4DefectOnt}};
    \draw[->] (defectont) -- (sensor);
    \draw[->] (sensor) -- (ssn);

  \end{tikzpicture}
	\caption{Diagram of the direct import structure for \emph{DefectOnt} modules.}
	\label{fig:mainclasses}	
\end{figure}
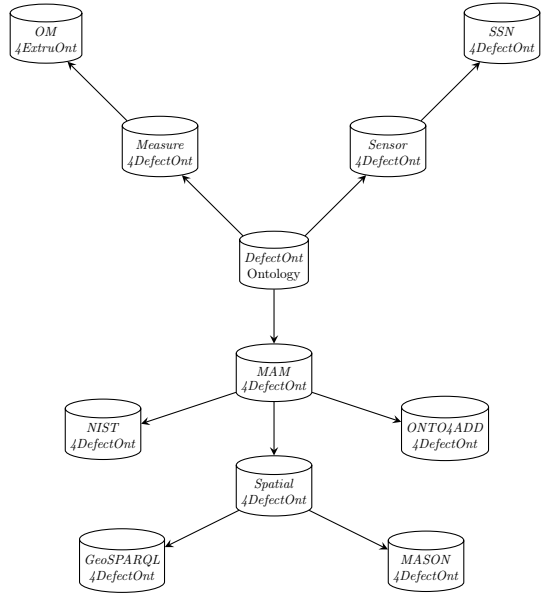

\noindent
Along the dimensions identified in the ontology development process (cf. Section~\ref{sec:ont-design-meth}) and depicted in Figure~\ref{fig:mainclasses}, the \DefectOnt modules provide relevant vocabulary and formalise the domain knowledge required to represent the main features of production defects possibly occurring in MAM processes.
We illustrate them in detail in the rest of this section, using DL syntax~\cite{BaaEtAl17} for the OWL 2 axioms here presented.

\subsection{\emph{MAM4DefectOnt} Module}
\label{sec:AM-mod}

\noindent
The \emph{MAM4DefectOnt} module is the primary module of \emph{DefectOnt} ontology. This module consists of
three
submodules:
\begin{enumerate*}[label=(\arabic*)]
	\item\label{item:nistmodule}\emph{NIST4DefectOnt},
	\item\label{item:ontomodule}\emph{ONTO4ADD4DefectOnt},
	and
	\item\label{item:spatialmodule}\emph{Spatial4DefectOnt}.
\end{enumerate*}
Modules~\ref{item:nistmodule} and~\ref{item:ontomodule} are obtained from the NIST AM ontology and the Onto4Additive ontology, respectively.
Module~\ref{item:spatialmodule} in turn consists of the following submodules:
\emph{GeoSPARQL4DefectOnt}, derived from GeoSPARQL1.1,
and
\emph{MASON4DefectOnt}, from
the MASON ontology.
In the following we present and discuss the relevant axioms of the \emph{MAM4DefectOnt} module, focusing on
the representation of
the different types and sources of MAM defects analysed in the literature.

\subsubsection{Defect types in \emph{MAM4DefectOnt}}
\label{sec:defecttype}
\begin{figure}[t]
	\includegraphics[width=\textwidth]{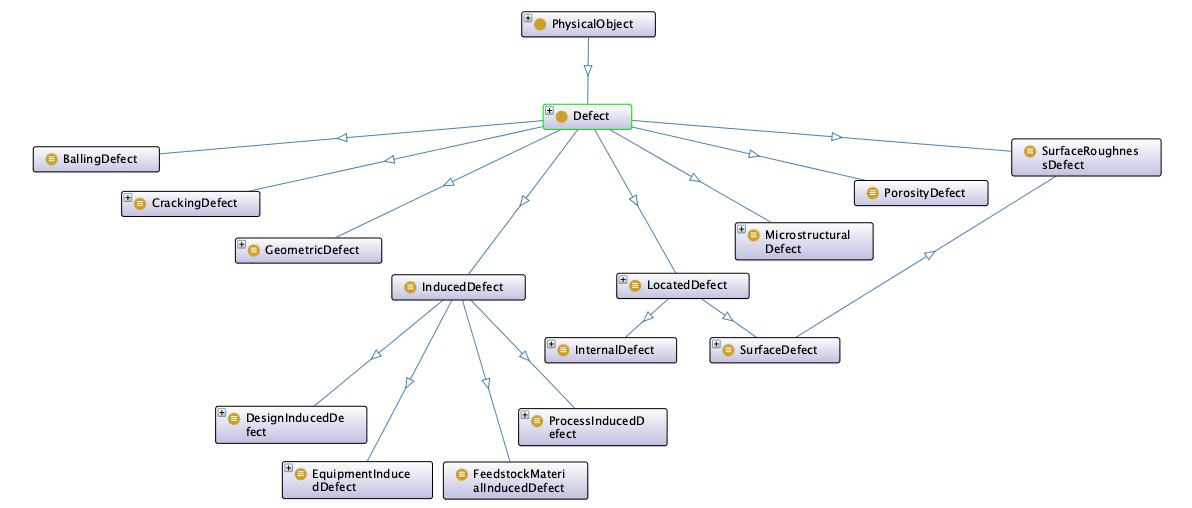}
	\caption{Main super- and sub-classes of the class ${\sf Defect}$ in \DefectOnt.}
	\label{fig:DefectOntomodules}	
\end{figure}

\noindent
The main class of the module used to represent the different kinds of MAM defects studied in the literature is the {\sf Defect} class (cf. Figure~\ref{fig:DefectOntomodules}).
On this class, we first impose the following constraint:
\begin{equation}
{\sf Defect} \sqsubseteq {\sf PhysicalObject} \sqcap \exists{\sf affects}.({\sf PhysicalArtefact} \sqcup {\sf Material}),
\end{equation}
where: {\sf PhysicalObject}, {\sf PhysicalArtefact}, and  {\sf Material} are classes in \emph{ONTO4ADD4DefectOnt}; {\sf affects} is a newly introduced object property, used to represent the relationship between a defect and a physical entity that is affected by it, having domain {\sf Defect}, range {\sf Physical} (from the \emph{NIST4DefectOnt} module), and inverse property {\sf isAffectedBy}.

As a further step, we leverage the \emph{Spatial4DefectOnt} module imported by \emph{MAM4DefectOnt} to provide a location-based classification of defects.
First, we define a \emph{located defect} as a defect that overlaps with some region of an AM layer, as follows:
\begin{equation}
	{\sf LocatedDefect} \equiv {\sf Defect} \sqcap \exists {\sf sfOverlaps}.{\sf AMLayerRegion},
\end{equation}
where:
{\sf sfOverlaps} comes from the \emph{Spatial4DefectOnt} submodule,
in turn obtained from GeoSPARQL1.1, where it represents the partially overlapping relation from the region connection calculus (RCC)~\cite{RanEtAl92};
and {\sf AMLayerRegion} is a newly introduced subclass of {\sf PhysicalRegion}, in turn set in \emph{MAM4DefectOnt} as subclass of {\sf PhysicalObject} from \emph{ONTO4ADD4DefectOnt}.
Moreover, we distinguish between \emph{internal defects} (i.e., defects located in overlapping with some infill regions, and with infill regions only, of an AM layer) and \emph{surface defects} (i.e., defects located so to overlap with contour regions of an AM layer).
The axioms formalising the above characterisation are the following:
\begin{align}	
	{\sf InternalDefect} \equiv \ & {\sf LocatedDefect} \sqcap \exists{\sf sfOverlaps}.{\sf InfillRegion} \sqcap \forall{\sf sfOverlaps}.{\sf InfillRegion}, \\
	{\sf SurfaceDefect} \equiv \ & {\sf LocatedDefect} \sqcap \exists{\sf sfOverlaps}.{\sf ContourRegion},
\end{align}
together with the axioms stating that 	{\sf ContourRegion} and {\sf InfillRegion}, together with {\sf PowderRegion}, are disjoint subclasses of {\sf AMLayerRegion}, and ${\sf AMLayer} \sqsubseteq \exists {\sf hasPart}.{\sf AMLayerRegion}$.
%
%
%
%

To formalise additional domain knowledge on MAM defects, we rely on the MAM defect classification proposed by Grasso and Colosimo in \cite{grasso2017process}. 
Due to its level of detail, this classification is preferred as the main reference framework over others in the AM literature (e.g., \cite{debroy2018additive,frazier2014metal,smith2016linking}). 
However, aiming at a broad and consistent representation model for MAM defects, we integrate in our ontology also other resources from the literature, adjusting where needed the overall structure of \citeauthor{grasso2017process} classification (cf. \autoref{sec:ont-design-meth}).
%

In \cite{grasso2017process}, the MAM defects are organised in six
main categories, represented in our module by corresponding subclasses of the ${\sf Defect}$ class:
\begin{enumerate*}[label=$(\roman*)$]
\item \emph{porosity} ({\sf PorosityDefect});
\item \emph{balling} ({\sf BallingDefect});
\item \emph{geometric defects} ({\sf GeometricDefect});
\item \emph{surface defects} ({\sf SurfaceRoughnessDefect});
\item \emph{cracks and delaminations} ({\sf CrackingDefect});
\item \emph{microstructural inhomogeneities and impurities} ({\sf MicrostructuralDefect}).
\end{enumerate*}
We introduce the {\sf SurfaceDefect} and {\sf InternalDefect} class to model purely location-based notions of defects, with respect to the AM layers they overlap with. Therefore, we employ the class {\sf SurfaceRoughnessDefect} to represent the specific kind of surface defects affecting the roughness of a product surface, that~\citeauthor{grasso2017process} simply refer to as \emph{surface defects}.

In order to present the ontology axioms used to connect these classes of defects with their corresponding sources, as well as to formalise their relationships with other AM concepts, we report in the following the main features of each group of defects.

\paragraph{Balling defects}
The surface tension generated in the melt pool can lead to melted balls of liquid metal that solidify into spherical particles, generating the so-called \textit{balling} phenomenon. These small spheres of metal may affect both the internal layers of the product, as well as its surface,
influencing the component surface roughness and porosity, and (if the phenomenon is very pronounced) it may eventually lead to (hemi-)spherical protrusion on the solidified structure. Balling can have dramatic effects on part quality and fatigue life of the component \cite{malekipour2018common}. 
To translate the above description into logical axioms, we distinguish between {\sf InternalBallingDefect} and {\sf SurfaceBallingDefect}
as subclasses covering
the {\sf BallingDefect} class, as follows:
%

\begin{align}
		\notag
		{\sf InternalBallingDefect} \equiv \
		\notag
		& {\sf InternalDefect} \ \sqcap \\
		\notag
		& \exists{\sf isConsequenceOf}.{\sf SurfaceTensionPhenomenon} \ \sqcap \\
		\notag
		& \exists{\sf isMadeOf}.({\sf Material} \sqcap \exists {\sf hasMaterialState}.\{ {\sf solidState} \})  \ \sqcap \\
		& \exists {\sf hasApproximateGeometry}.{\sf Ball}, \\
		\notag
		{\sf SurfaceBallingDefect} \equiv \
		& {\sf SurfaceDefect} \ \sqcap \\
		\notag
		& \exists{\sf isConsequenceOf}.{\sf SurfaceTensionPhenomenon} \ \sqcap \\
		\notag
		& \exists{\sf isMadeOf}.({\sf Material} \sqcap \exists {\sf hasMaterialState}.\{ {\sf solidState} \})  \ \sqcap \\
		& \exists {\sf hasApproximateGeometry}.{\sf HemiBall}, \\
	{\sf BallingDefect} \equiv \ &
	{\sf InternalBallingDefect} \sqcup {\sf SurfaceBallingDefect},
\end{align}
where:
{\sf isConsequenceOf} is a property introduced in our \emph{MAM4DefectOnt} module;
the property {\sf hasApproximateGeometry}, as well as the classes {\sf Ball} and {\sf HemiBall}, are introduced in \emph{Spatial4DefectOnt};
 {\sf SurfaceTensionPhenomenon} is set in \emph{MAM4DefectOnt}  as a subclass of {\sf Phenomenon}, in turn inherited from the \emph{NIST4DefectOnt} submodule;
{\sf isMadeOf} is an object property of \emph{MASON4DefectOnt};
while {\sf Material}, {\sf hasMaterialState}, and {\sf solidState} are, respectively, a class, a property and an individual taken from \emph{ONTO4ADD4DefectOnt}.
%

\paragraph{Cracking and delamination defects}
According to \cite{debroy2018additive} there are three main cracking mechanisms observed in MAM: $(i)$ solidification cracking along the boundaries, $(ii)$ liquation cracking occurring either in the mushy region or in the partially melted zone, and $(iii)$ delamination, consisting in the separation of two consecutive printed layers. All the three types of cracking are related to the high residual stresses induced by the process \cite{mercelis2006residual} as a consequence of stress relief through fracture. This characterising feature of cracking defects is formalised in our ontology as follows:
\begin{align}
\notag
{\sf CrackingDefect} \equiv \ & {\sf Defect} \ \sqcap \\
\notag
& \exists{\sf affects}.({\sf AMProduct} \ \sqcap \\
& \exists{\sf hasFeature}.(\exists{\sf isConsequenceOf}.{\sf Fracture}))
\end{align}
with: {\sf hasFeature} object property of \emph{ONTO4ADD4DefectOnt};
and class {\sf Fracture} introduced in \emph{MAM4DefectOnt} within the subclass hierarchy of {\sf Phenomenon} from \emph{NIST4DefectOnt}.
In our module, we also introduce specific subclasses for each type of cracking, in order to better specialise their relationship with the process, namely: {\sf SolidificationCrackingDefect}, {\sf LiquationCrackingDefect}, and {\sf DelaminationCrackingDefect}.
%
%

\paragraph{Geometric defects}
This class of defects includes all the different kinds of dimensional and geometric deviations from the original, as-designed geometry \cite{malekipour2018common}.
In fact, it is widely documented in the literature that complex, lattice, and slender 3D printed structures present non negligible geometric deviation from the the actual, as-build structure and the as-design geometry as generated within Computer Aided Design (CAD) environments \cite{korshunova2021bending,BONIOTTI2019105181,KORSHUNOVA2021101949}. Such kind of defects is defined in the \emph{MAM4DefectOnt} module, as follows:
\begin{align}
		\notag
		{\sf GeometricDefect} \equiv \ & {\sf Defect} \ \sqcap \\
		\notag
		& \exists {\sf affects}.({\sf AMProduct} \sqcap \exists{\sf hasGeometry}.({\sf AsBuiltGeometry} \ \sqcap \\
		& \exists{\sf hasSignedDeviationFrom}.{\sf AsDesignedGeometry})),
\end{align}
%
where:
{\sf hasGeometry} and {\sf hasSignedDeviationFrom} are properties, and {\sf AsBuiltGeometry} and {\sf AsDesignedGeometry} are classes introduced in the \emph{Spatial4DefectOnt} module.

The most common effects leading to geometric defects are: $(i)$ shrinkage, $(ii)$ warping, $(iii)$ curling, and $(iv)$ formation of super-elevated edges. They are inserted into the module as subclasses of {\sf GeometricDefect} and defined as: {\sf ShrinkageGeometricDefect}, {\sf WarpingGeometricDefect}, {\sf CurlingGeometricDefect}, {\sf SuperelevatedEdgesGeometricDefect}, respectively.
\paragraph{Microstructural defects}
Several different classifications can be found in MAM literature to label and classify the defects related to the microstructure of 3D printed components \cite{MIKULA2021109851,song2015differences}. 
For our purposes, to provide an axiom characterising microstructural defects, we first introduce in \emph{MAM4DefectOnt} the class {\sf InhomogeneousMicrostructure} as a subclass of {\sf Microstructure}, that we inherit from the
NIST ontology.
The axiom formalising that a microstructural defect is any defect that affects an AM product with inhomogeneous microstructure is then written as follows:
\begin{align}
\notag
	{\sf MicrostructuralDefect} \equiv \ & {\sf Defect} \sqcap \exists{\sf affects}.({\sf AMProduct} \ \sqcap \\
	& \exists{\sf presents}.{\sf InhomogeneousMicrostructure})
\end{align}
%
where
{\sf presents} is a newly introduced object property in \emph{MAM4DefectOnt}, used to relate physical objects with the characteristics they exhibit.
Finally, following \citet{sharratt2015non}, we distinguish between three types of inhomogeneities of the microstructure that occur during MAM processes, introducing them as corresponding subclasses of {\sf MicrostructuralDefect}: $(i)$ impurities ({\sf ImpuritiesMicrostructuralDefect}), $(ii)$ grain size characteristics ({\sf GrainSizesMicrostructuralDefect}), and $(iii)$ crystallographic textures ({\sf CrystallographicTexturesMicrostructuralDefect}).
%

\begin{figure}[h!]
	\centering
	\subfloat[Internal porosity defect.\label{subfig:smallPores}]
	{
		\includegraphics[width=0.458\textwidth]{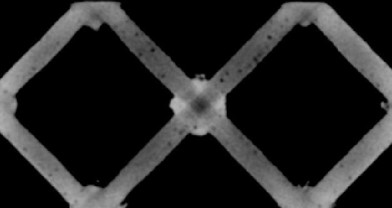}
	}
	\subfloat[Surface porosity defect.\label{subfig:largePores}]
	{
		\includegraphics[width=0.4\textwidth]{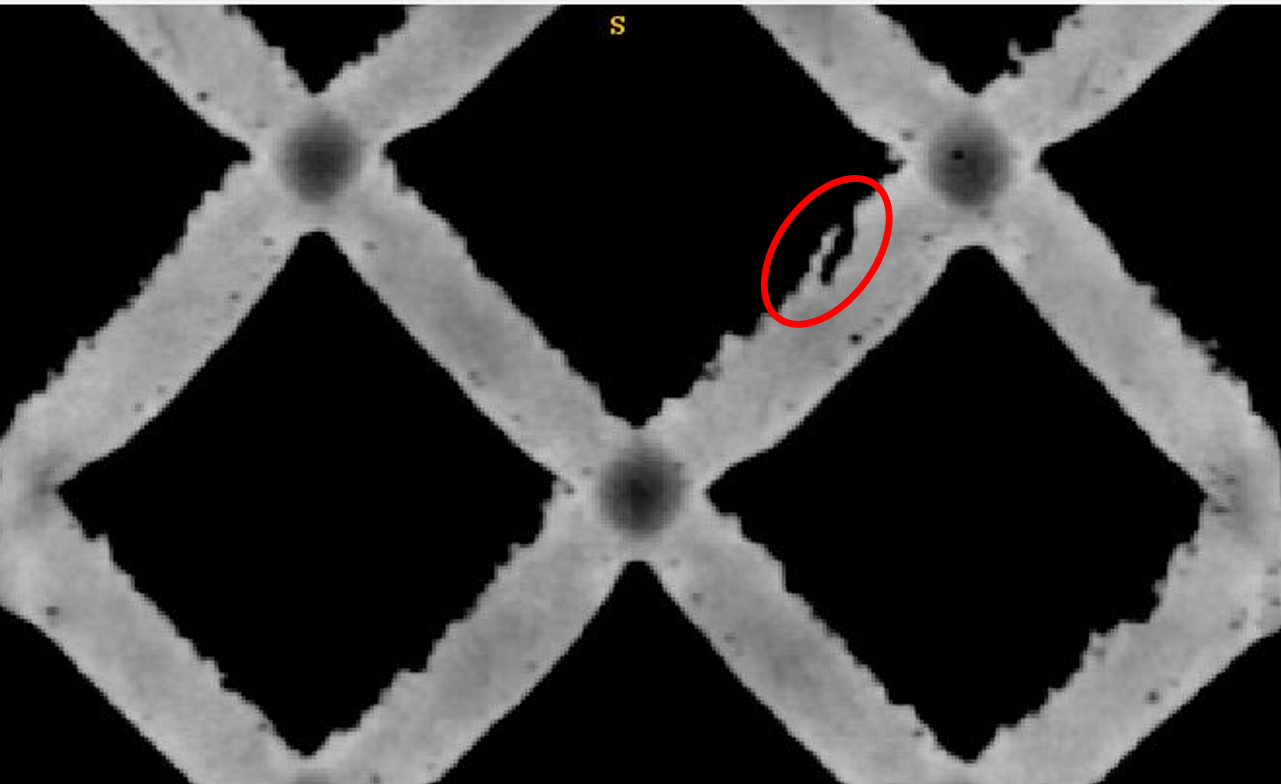}
	}    
	\caption{View by micro-Computed Tomography showing different kinds of pores in a Stainless Steel 316L 3D printed specimen~\cite{carraturo2021experimental}.}
	\label{fig:poresPic}
\end{figure}

\paragraph{Porosity defects}
Fully dense (99.95$\%$) MAM parts can be produced when printing large bulk material components. However, more often a functional design and/or lightweight structures (e.g., lattice) are present in MAM products (cf.~\cref{fig:poresPic}). In this case porosity -- due, e.g, to lack-of-fusion or keyhole effects -- can severely affect the performances of 3D printed components. Based on the location of the defects, 
we further specify the notions of \emph{internal} and \emph{surface} porosity defects (cf. Figure~\ref{subfig:smallPores} and~\ref{subfig:largePores}, respectively), formalising them respectively as {\sf InternalPorosityDefect} and {\sf SurfacePorosityDefect},
as follows:
\begin{align}	
		{\sf InternalPorosityDefect} \equiv \ & {\sf InternalDefect} \sqcap {\sf VoidRegion} \sqcap \exists{\sf hasBoundary}.{\sf ClosedSurface} \\
		{\sf SurfacePorosityDefect} \equiv \ & {\sf SurfaceDefect} \sqcap {\sf VoidRegion} \\
		{\sf PorosityDefect} \equiv \ & {\sf SurfacePorosityDefect} \sqcup {\sf InternalPorosityDefect}
\end{align}
%
where:
{\sf isBoundaryOf} and {\sf ClosedSurface} are, respectively, a property and a class introduced in \emph{Spatial4DefectOnt};
${\sf VoidRegion}$ is a newly introduced class (cyclically) defined in \emph{MAM4DefectOnt} as
${\sf VoidRegion} \equiv {\sf PhysicalObject} \ \sqcap \  \forall {\sf hasPart}.{\sf VoidRegion}$.


\paragraph{Surface roughness defects}
Surface roughness defects are clearly related to the notion of surface roughness, i.e., to a deviation of the surface texture with respect to an ideal surface plane. In MAM processes, surface roughness is influenced by two main effects: the stair-stepping effects, due to the layer-by-layer nature of the process, and the actual roughness of the metal surface \cite{fox2016,strano2013surface}. Additionally, the relative position of the surface strongly influences its roughness, e.g., downward surfaces present much higher roughness than upward surfaces.
We formalise the connection between surface roughness defects and the rough surface texture of a product, as follows:
\begin{align}
		\notag
		{\sf SurfaceRoughnessDefect} \equiv \ & {\sf SurfaceDefect}  \ \sqcap \\
		& \exists {\sf affects}.({\sf AMProduct} \sqcap
		\exists{\sf presents}.{\sf RoughSurfaceTexture})
\end{align}
where:
{\sf AMProduct} is a class from the \emph{ONTO4ADD4DefectOnt} module; and {\sf RoughSurfaceTexture} is introduced as subclass of {\sf SurfaceTexture} in \emph{MAM4DefectOnt}.


%
\subsubsection{Defect sources in \emph{MAM4DefectOnt}}
\label{sec:defectsource}

\noindent
In order to establish logical connections between MAM defects and their causes,
a further classification of defects based on the corresponding sources is required.
To this goal, we introduce {\sf InducedDefect} as a subclass of {\sf Defect}, defining it as follows:
\begin{align}	
\notag
		{\sf InducedDefect} \equiv \ & {\sf Defect} \sqcap \exists{\sf isInducedBy}.({\sf Characteristic} \sqcup  {\sf Material} \ \sqcup \\
		& {\sf Phenomenon} \sqcup {\sf PhysicalObject} \sqcup  {\sf Process})
\end{align}
where the newly introduced object property {\sf isInducedBy} represents the relationship between defects and those MAM entities that are known to have an influence on them.
According to the classification proposed by \citet{grasso2017process}, Table~1 (and references therein), that we further extend and specify,
MAM defect sources are classified into four main groups, for which we introduce a corresponding  subclass of {\sf InducedDefect}:
\begin{enumerate*}[label=$(\roman*)$]
	\item \emph{design for additive choices} ({\sf DesignInducedDefect});
	\item \emph{equipment} ({\sf EquipmentInducedDefect});
	\item \emph{feedstock material} ({\sf FeedstockMaterialInducedDefect}); and 
	\item \emph{process} ({\sf ProcessInducedDefect}). 
\end{enumerate*}
%
In the following, we discuss the axioms that characterise each of the {\sf InducedDefect} subclasses.

%
\paragraph{Design-induced defects}
A wrong support design and/or part orientation during the building process can have a dramatic impact on the part quality and performances of 3D printed components. In fact, these two parameters directly influence $(i)$ dimensional accuracy, $(ii)$ surface roughness, and $(iii)$ part microstructure \cite{kleszczynski2012error,foster2015optical,zeng2015optimization,strano2013surface}.
Accordingly, in the \emph{MAM4DefectOnt} module the class {\sf DesignInducedDefect} includes two other classes, {\sf SupportsInducedDefect} and {\sf OrientationInducedDefect}, defined by the following axioms:
\begin{align}	
		{\sf SupportsInducedDefect} \equiv \ & {\sf DesignInducedDefect} \sqcap \exists{\sf isInducedBy}.{\sf AMSupportStructure} 
\\
	{\sf OrientationInducedDefect} \equiv \ & {\sf DesignInducedDefect} \sqcap \exists{\sf isInducedBy}.{\sf ProductOrientation}
\end{align}
where {\sf AMSupportStructure} comes from \emph{ONTO4DD4DefectOnt}, while {\sf ProductOrientation} is a new subclass of {\sf ProductCharacteristic} introduced in \emph{MAM4DefectOnt}.

\paragraph{Equipment-induced defects}
Improper or sub-optimal setting and calibration of the MAM equipment can generate several types of defects. This category includes four main sources of defects:
$(i)$ beam scanning and deflection system;
$(ii)$ powder handling and deposition system;
$(iii)$ insufficient baseplate thickness;
and 
$(iv)$ build chamber environmental control (e.g., preheating of the powder bed) \cite{foster2015optical,kleszczynski2012error,prabhakar2015computational,spears2016process}. Each of these sources can be linked to different types of defects. For instance, an erroneous calibration of the beam scanning can lead to longer exposure time distorting the melt pool morphology, which directly affects residual stresses, geometric accuracy, and porosity (lack of fusion defects) in the final component \cite{yeung2016,marques2020numerical,vastola2016modeling,carraturo2020numerical}.
In the \emph{MAM4DefectOnt} module, we introduce the class {\sf EquipmentInducedDefect} that, in accordance with the observations above, includes four subclasses: {\sf BaseplateInducedDefect}, {\sf BeamScanningDeflectionSystemInducedDefect}, {\sf BuildChamberEnvironmentalControlInducedDefect}, and {\sf PowderHandlingDepositionSystemInducedDefect}.
In turn, these are defined by the following axioms:
%
\begin{align}	
		\notag
		{\sf EquipmentInducedDefect} \equiv \ & {\sf InducedDefect} \ \sqcap \ \\
		\notag
		& \exists{\sf isInducedBy}.({\sf EquipmentParameter} \ \sqcup \\
		\notag
		& {\sf MfgDevice} \ \sqcup \\
		& ({\sf Material} \sqcap \exists {\sf makes}.{\sf MfgDevice})) \\
		\notag
		{\sf BaseplateInducedDefect} \equiv \ & {\sf EquipmentInducedDefect} \ \sqcap \ \\
		\notag
		& \exists{\sf isInducedBy}.({\sf BaseplateMaterial} \ \sqcup  \\
		\notag
		& {\sf BaseplatePreHeatingTemperature} \ \sqcup \\
		& {\sf BaseplateThickness}) \\
		\notag
		{\sf BeamScanningDeflectionSystem} \phantom{\ \equiv \ } & \\
		\notag
		{\sf InducedDefect} \equiv \ & {\sf EquipmentInducedDefect} \ \sqcap \\
		& \exists{\sf isInducedBy}.{\sf OpticalParameter}	\\
		\notag
		{\sf BuildChamberEnvironmentalControl} \phantom{\ \equiv \ } & \\
		\notag
		{\sf InducedDefect} \equiv \ & {\sf EquipmentInducedDefect} \	\sqcap \\
		& \exists{\sf isInducedBy}.{\sf BuildChamberParameter} \\
		\notag
		{\sf PowderHandlingDepositionSystem} \phantom{\ \equiv \ } & \\
		\notag
		{\sf InducedDefect} \equiv \ & {\sf EquipmentInducedDefect} \ \sqcap \\
		\notag
		& \exists{\sf isInducedBy}.({\sf MaterialDepositionSystem} \ \sqcap \\
		\notag
		& \exists{\sf isAffectedBy}.({\sf ByproductMaterialEjectionInducedDefect} \ \sqcup \\
		& {\sf SuperelevatedEdgesGeometricDefect}))
\end{align}
%
It might be noted that defects induced by the powder handling and deposition system ({\sf PowderHandlingDepositionSystemInducedDefect}) are not influenced by recoating system characteristics (e.g., the wear of the recoater). In fact, most of the time the recoating system is not the direct source of these defects \cite{kleszczynski2012error,foster2015optical}. Instead, they are generated by the linear motion of the recoater that propagates inhomogeneities present on the powder bed (due to, e.g., super-elevated edges and spatters) along its moving direction.

\paragraph{Feedstock material-induced defects}
Powder quality, i.e., powder morphology, directly influences the quality (geometrical accuracy, porosity, and microstructure) as well as the performances (fatigue life) of the component \cite{sames2016metallurgy,das2003physical,van2007complexity,niu1999instability}. Another important feature of the powder is its flowability (powder fluidity), which depends on the particle size. If the powder particles are too small or too big, a smooth layer deposition cannot be achieved, eventually even stopping the process. Moreover, uniform powder deposition, smooth deposited powder layer, and uniform spreading, affect surface quality \cite{malekipour2018common,seyda2012investigation}.
Based on the observations above, we define the class {\sf FeedstockMaterialInducedDefect} by means of the following axiom:
\begin{align}
\notag
{\sf FeedstockMaterialInducedDefect} \equiv \ & {\sf InducedDefect} \ \sqcap \\
& \exists{\sf isInducedBy}.({\sf PackingDensity} \sqcup {\sf PowderSizeDistribution})
\end{align}

\paragraph{Process-induced defects}
Process parameters (e.g., laser power, scan speed, and hatch distance) influence the energy density that is introduced into the powder bed domain \cite{khairallah2016laser}. The energy density has been demonstrated to control the melt pool morphology and the cooling rate, thus ultimately influencing material microstructure \cite{LIAN2019107672}, geometric accuracy \cite{yeung2018}, as well as the porosity of final components \cite{MIKULA2021109851}.
Therefore, we include in the \emph{MAM4DefectOnt} module the class {\sf ProcessInducedDefect}, together with the two subclasses {\sf ByproductsAndMaterialEjectionInducedDefect} and {\sf ParameterAndScanStartegyInducedDefect} defined as follows:
\begin{align}
\notag
	{\sf ProcessInducedDefect} \equiv \ & {\sf InducedDefect} \ \sqcap
	\notag
	 \exists {\sf isInducedBy}.
({\sf Process} \ \sqcup \\
	& {\sf ProcessParameter}) \\
		{\sf ByproductMaterialEjectionInducedDefect} \equiv \ & {\sf InducedDefect} \sqcap \exists{\sf isInducedBy}.{\sf Ejection} \\
		\notag
		{\sf ParameterScanStrategyInducedDefect} \equiv \ & {\sf InducedDefect} \ \sqcap \\
		\notag
		& \exists{\sf isInducedBy}.({\sf HatchingDistance} \ \sqcup \\
		\notag
		& {\sf HeatSourcePower} \ \sqcup \\
		& {\sf LayerThickness} \sqcup {\sf ScanningSpeed})
\end{align}

\subsubsection{Relating defects types and sources in \emph{MAM4DefectOnt}}
\label{sec:bridgeaxioms}

\noindent
Following~\citet{grasso2017process}, Table~1 (and references therein), in this section we discuss the mappings connecting each type of MAM defects formalised in Section~\ref{sec:defecttype} with the corresponding causes identified in the literature, and modelled in terms of {\sf InducedDefect} subclasses in Section~\ref{sec:defectsource}.
Such mappings consist of subsumption axioms of the form
$A \sqsubseteq C$
where: $A$ is any of the classes {\sf BallingDefect}, {\sf CrackingDefect}, {\sf GeometricDefect}, {\sf MicrostructuralDefect}, {\sf PorosityDefect}, {\sf SourfaceRoughnessDefect}; and $C$ is the union of subclasses of {\sf InducedDefect}.
These axioms
express the fact that defects of type $A$ are
documented
in the literature to be caused by at least one of the MAM features that characterise some disjoints in $C$.

For instance, it is
documented in
the literature~\cite{grasso2017process} that porosity defects can be induced by:
the beam scanning or deflection system;
the build chamber environmental control;
the powder handling and deposition system;
the process parameters and scan strategies;
byproducts and material ejections.
Accordingly, we express in \emph{MAM4DefectOnt} that the class {\sf PorosityDefect} is included in the union of
the following classes:
{\sf BeamScanningDeflectionSysemInducedDefect},
{\sf BuildChamberEnvironmentalControlInducedDefect},
{\sf PowderHandlingDepositionSystemInducedDefect},
{\sf ParameterScanStrategyInducedDefect},
{\sf ByproductMaterialEjectionInducedDefect},
and
{\sf FeedstockMaterialInducedDefect}.

In the rest of this section, we list the mappings axioms for each of the subclasses of {\sf Defect} discussed in Section~\ref{sec:defecttype}.

\paragraph{Balling defect sources}

\begin{align}
		\notag
	{\sf BallingDefect} \sqsubseteq \ & {\sf BuildChamberEnvironmentalControlInducedDefect} \ \sqcup \\
		\notag
		& {\sf ParameterScanStrategyInducedDefect} \ \sqcup \\
		& {\sf OrientationInducedDefect}
\end{align}


\paragraph{Cracking and delamination defect sources}

\begin{align}
		\notag
	{\sf CrackingDefect} \sqsubseteq \ & {\sf BuildChamberEnvironmentalControlInducedDefect} \ \sqcup \\
		\notag
		& {\sf BaseplateInducedDefect} \ \sqcup \\
		\notag
		& {\sf ParameterScanStrategyInducedDefect} \ \sqcup \\
		& {\sf SupportInducedDefect}
\end{align}


\paragraph{Geometric defect sources}

\begin{align}
		\notag
	{\sf GeometricDefect} \sqsubseteq \ & {\sf BeamScanningDeflectionSystemInducedDefect} \ \sqcup \\
		\notag
		& {\sf PowderHandlingDepositionSystemInducedDefect} \ \sqcup \\
		\notag
		& {\sf BaseplateInducedDefect} \ \sqcup \\
		\notag
		& {\sf ParameterScanStrategyInducedDefect} \ \sqcup \\
		\notag
		& {\sf SupportInducedDefect} \ \sqcup \\
		\notag
		& {\sf OrientationInducedDefect} \ \sqcup \\
		& {\sf FeedstockMaterialInducedDefect}
\end{align}


\paragraph{Microstructural defect sources}

\begin{align}
		\notag
	{\sf MicrostructuralDefect} \sqsubseteq \ & {\sf BuildChamberEnvironmentalControlInducedDefect} \ \sqcup \\
			\notag
		& {\sf PowderHandlingDepositionSystemInducedDefect} \ \sqcup \\
		\notag
		& {\sf ParameterScanStrategyInducedDefect} \ \sqcup \\
		\notag
		& {\sf ByproductMaterialEjectionInducedDefect} \ \sqcup \\
		\notag
		& {\sf OrientationInducedDefect} \ \sqcup \\
		& {\sf FeedstockMaterialInducedDefect}
\end{align}


\paragraph{Porosity defect sources}

\begin{align}
		\notag
	{\sf PorosityDefect} \sqsubseteq \ & {\sf BeamScanningDeflectionSystemInducedDefect} \ \sqcup \\
		\notag
		& {\sf BuildChamberEnvironmentalControlInducedDefect} \ \sqcup \\
		\notag
		& {\sf PowderHandlingDepositionSystemInducedDefect} \ \sqcup \\
		\notag
		& {\sf ParameterScanStrategyInducedDefect} \ \sqcup \\
		\notag
		& {\sf ByproductMaterialEjectionInducedDefect} \ \sqcup \\
		& {\sf FeedstockMaterialInducedDefect}
\end{align}


\paragraph{Surface roughness defect sources}

\begin{align}
		\notag
	{\sf SurfaceRoughnessDefect} \sqsubseteq \ 
		& {\sf PowderHandlingDepositionSystemInducedDefect} \ \sqcup \\
		\notag
		& {\sf ParameterScanStrategyInducedDefect} \ \sqcup \\
		\notag
		& {\sf SupportInducedDefect} \ \sqcup \\
		\notag
		& {\sf OrientationInducedDefect} \ \sqcup \\
		& {\sf FeedstockMaterialInducedDefect}
\end{align}


\subsubsection{Competency questions for \emph{MAM4DefectOnt}}
\label{sec:CQsMAM4DefectOnt}
\noindent
The \emph{MAM4DefectOnt} module, together with its \emph{Spatial4DefectOnt} submodule, allow us to answer, for instance, to the following CQs:
%
				%
				\begin{enumerate}[label={CQAm.\arabic*}, leftmargin=*]
					\item[CQAm.2] Is the feature ${\sf d}$ an instance of a porosity defect?
				\end{enumerate}
				\begin{enumerate}[label={CQSp.\arabic*}, leftmargin=*]
					\item[CQSp.2] What is the length in metres of the cracking defect instance {\sf d}?
				\end{enumerate}

\subsection{\emph{Measure4DefectOnt} Module}
\label{sec:meas-mod}

\noindent
This module is intended to provide axioms, classes, and properties that are necessary to describe AM process and product features requiring notions related to measures, measuring units, and scales. Relevant classes (e.g., {\sf Temperature}, {\sf CelsiusTemperature}, {\sf Area}, {\sf Length}) are included in its only submodule, \emph{OM4ExtruOnt}, which is  imported from the \emph{ExtruOnt}~\cite{RamEtAl20} by leveraging the modular structures of both \emph{ExtruOnt} and \emph{DefectOnt} ontologies. 
\citeauthor{RamEtAl20} obtain this submodule from the OM ontology via a pruning process that retains only those concepts and properties pertaining to manufacturing process and components.
In particular, this module is intended to provide the necessary knowledge to address, for instance, the following competency question:
%
\begin{enumerate}[label={CQMe.\arabic*}, leftmargin=*]
	\item[CQMe.3] What is the length in meters of the cracking defect instance {\sf d}?
\end{enumerate}
\subsection{\emph{Sensor4DefectOnt} Module}
\label{sec:sens-mod}
This module is intended to enable domain experts and/or online defect detection systems to gain insights from data measured by means of sensors installed on the AM machine. In particular, the axioms of the class {\sf FeatureOfInterest} together with its related properties and subclasses are present in this module through the submodule of the SSN ontology, \emph{SSN4DefectOnt} .
Moreover, this module provides useful concepts to evaluate the quality of MAM processes allowing to potentially perform online defect diagnosis and troubleshooting. In particular, \emph{sensor4DefectOnt} allows to answer the following competency questions:
\begin{enumerate}[label={CQSe.\arabic*}, leftmargin=*]
					\item[CQSe.2] Does any sensor hosted by platform {\sf pl} observe an instance of a porosity defect?
\end{enumerate}
 


%
\section{Conclusion}\label{sec:conclusions}
\noindent
In the present contribution, we have introduced an OWL 2 ontology to represent MAM defects, together with their constituting features and sources. Our ontology aims at filling a gap in the literature on AM knowledge bases, providing dedicated vocabulary and axioms to capture specific defect-related domain knowledge. To provide additional expressivity and modelling capabilities, we have also integrated our ontology with several upper-level ontologies, as well as domain-specific ontologies from different domains.

As future steps, we plan to extend \emph{DefectOnt} with the capability of relating defects with measured quantities coming from either \textit{in-situ} or \textit{ex-situ} measurements.
Moreover, we are interested in improving the alignment of \DefectOnt with upper ontology or domain-specific knowledge bases from the AM literature.
In a similar direction, to foster the integration of heterogeneous data sources, we plan to consider ontology-based data access techniques and frameworks, focusing in particular on the OWL 2 QL Profile~\cite{KonEtAl13}.
Finally, we are interested in investigating the addition of a temporal dimension to our ontology, for temporal data modelling and predictive analysis~\cite{ArtEtAl13,BaaEtAl13,BraEtAl17,KalEtAl19}

%
Our long term goal is to use these additional representation and reasoning features to mitigate manufacturing defects, by automatically adjusting those process parameters that are found to be related with the corresponding defect sources.
Indeed, our defect ontology is meant to pave the way towards real-time control systems for MAM processes. In fact, combining machine learning mechanisms with knowledge-based representation and high-level reasoning techniques, it can potentially lead to a self-adaptive system able to perform online diagnosis and troubleshooting. 
%

%
\bibliographystyle{plainnat}          
\bibliography{bibliography}        
\end{document}